%% file: main.tex
\documentclass{article}
\usepackage{graphicx} 
\usepackage{tabularx}
\usepackage{amsmath}
\usepackage{xspace}
\usepackage{booktabs}
\usepackage{tcolorbox}

\usepackage{colortbl}
\usepackage{graphicx}
\usepackage{enumitem}

\usepackage{wrapfig}

\usepackage{amsmath}
\usepackage{amsfonts}
\usepackage{bm}
\usepackage{vcell}
\usepackage{hyperref}

\usepackage[font=scriptsize,labelfont=sl,labelsep=period]{caption}

\usepackage[preprint]{corl_2023} % Uncomment for pre-prints (e.g., arxiv); This is like ``final'', but will remove the CORL footnote.

\newcommand{\system}[0]{\textsc{AR2-D2}\xspace}
\newcommand{\model}[0]{\textsc{PerAct}\xspace}

\title{\includegraphics[height=7mm]{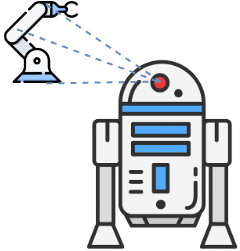}
AR2-D2:\\ Training a Robot Without a Robot}

\author{ Jiafei Duan~$^{1,}$
 \hspace{4px} Yi Ru Wang~$^1$ \hspace{4px} Mohit Shridhar~$^1$ \hspace{4px} Dieter Fox~$^{1, 2}$\hspace{4px} Ranjay Krishna~$^{1, 3}$\\
$^1$University of Washington \hspace{6px} $^2$NVIDIA \hspace{6px} $^3$Allen Institute for Artificial Intelligence\\
\tt\small \{duanj1, yiruwang, mshr, fox, ranjay\}@cs.washington.edu\\[1em]
\large\textbf{\url{www.ar2d2.site}}
}
\begin{document}
\maketitle

\begin{abstract}
\input{sections/0_abstract}
\end{abstract}


\input{sections/1_introduction}
\input{sections/2_related_work}
\input{sections/3_method}

\input{sections/4_user_study}

\input{sections/5_experiments}
\input{sections/6_conclusion}

% \clearpage
% \acknowledgments{}

%===============================================================================
\bibliography{example}

\end{document}

%% file: sections/0_abstract.tex
Diligently gathered human demonstrations serve as the unsung heroes empowering the progression of robot learning.
Today, demonstrations are collected by training people to use specialized controllers, which (tele-)operate robots to manipulate a small number of objects.
By contrast, we introduce \system: a system for collecting demonstrations which (1) does not require people with specialized training, (2) does not require any real robots during data collection, and therefore, (3) enables manipulation of diverse objects with a real robot.
\system\ is a framework in the form of an iOS app that people can use to record a video of themselves manipulating any object while simultaneously capturing essential data modalities for training a real robot. We show that data collected via our system enables the training of behavior cloning agents in manipulating real objects.
Our experiments further show that training with our AR data is as effective as training with real-world robot demonstrations.
Moreover, our user study indicates that users find \system\ intuitive to use and require no training in contrast to four other frequently employed methods for collecting robot demonstrations. 

% Painstakingly-collected human demonstrations are the inglorious heroes enabling robot learning.
% Today, demonstrations are collected by training people to use specialized controllers, which (tele-)operate robots to manipulate a small number of objects.
% By contrast, we introduce \system: a system for collecting demonstrations (1) without the need to train people, (2) without any real robots during demonstration collection, and therefore, (3) enabling manipulation of diverse objects.
% \system\ is an iOS app that people can use to record a video of themselves manipulating any object whilst simultaneously capturing the essential data modalities for training a real robot. We show that data collected via our system  enables the training of behavior cloning agents in manipulating real objects.
% Our experiments further show that training with \system's AR data is as effective as training with real-world demonstrations.
% Moreover, our user study indicates that users find \system\ intuitive to use and require no training compared to five other collection methods; users also produce demonstrations as fast as kinesthetic-teaching.

%% file: sections/1_introduction.tex
\section{Introduction}

% MAIN RESULTS:
% 1: system
% 2: robot can learn to manipulate objects that have never been interacted with a real robot. 
% 3: can now train a robot to interact with a personalized objects that is not obtainable en masse
% 4: manipulation results are as successful as those collected using real demonostrations with a real robot
% 5: users find it easy to use, as fast as kinesthetic data collection, human learning curve is faster, no training
% minor ablations: optimal number of demonstrations for finetuning, rbg-only variants, learning from simulation.

% Overall layout of the introduction.
% Para 1: robot data collection is really vital and powers a lot of machine learning, especially in robotics
Manually curated datasets are often the inglorious heroes of many large-scale machine learning projects~\cite{radford2021learning,ramesh2022hierarchical, duan2022survey}; this is especially true in robotics, where human-generated datasets of robot demonstrations are indispensable~\cite{dasari2019robonet,ebert2021bridge} especially with recent success in robot learning via imitation learning \cite{shridhar2022perceiver,wang2023mimicplay,ahn2022can,jiang2022vima} of these demonstration data.
%Since pure reinforcement learning requires an immense amount of simulations for robot agents to uncover useful behavior, behavior cloning from human demonstrations has become the de-facto training mechanism to initialize agents~\cite{duan2017one}.
For example, one recent effort collected $\sim130k$ robot demonstrations, with a fleet of $13$ robots over the course of $17$ months~\cite{brohan2022rt}.
As a result, researchers have spent considerable effort developing various mechanisms for demonstration collection. 
One popular option for collecting robot demonstrations is through kinesthetic-teaching, where a person guides a robot through a desired trajectory~\cite{argall2009survey}. Although intuitive, this mechanism can be tedious and slow~\cite{osentoski2010crowdsourcing}. Alternatively, teleoperation with various controllers has become popular: using a keyboard and mouse~\cite{kent2017comparison,leeper2012strategies}, a video game controller~\cite{laskey2017comparing}, a 3D-mouse~\cite{dragan2012online,shridhar2022perceiver}, special purpose master-slave interfaces~\cite{akgun2012novel,liang2017using}, and even virtual reality (VR) controllers~\cite{whitney2019comparing,zhang2018deep,lipton2017baxter}.

% Para 2: robot data collection is really challenging because people need to be trained to use methods to operate real robots; on top of that, data collection is limited by how bulky robots are: they are stuck in `warehouses'. To make matters worse, robots are expensive and demonstration collection is limited by the number of available robots.
Despite all these demonstration collection efforts, there are three key challenges limiting robot data collection. First, people need to be trained to produce useful demonstrations: kinesthetic methods are labor-intensive while teleoperation methods require learning specialized controllers. 
Second, the ability to parallelize data collection is limited by how many—often expensive—robots are available. Third, robots are usually bulky and locked within a laboratory, reducing their exposure to a handful of nearby objects. Lastly, (tele)-operation in simulation has the potential to scale efficiently without real robot hardware, but addressing the sim2real gap and limited variety of trainable tasks in simulation environments are challenges to overcome.

% Para 3: To make robot demonstrations collection effortless, we introduce \system. Describe what it is.

\begin{figure}[t]
    \centering
    \includegraphics[width=\textwidth]{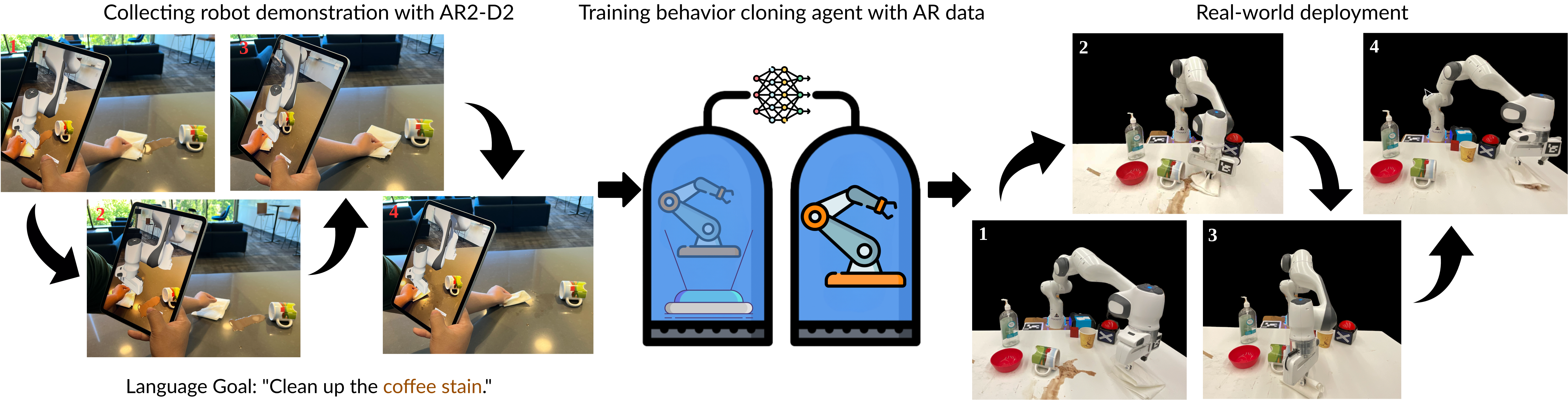}
    \caption{\textbf{\system\ } collects robot demonstrations without needing a real robot. \textbf{(Left)} Using \system, the user captures a video manipulating an object with their arm. \system projects an operational URDF of an AR Franka Panda robot arm into a physical environment. It uses a hand-pose tracking algorithm to move the AR robot's end effector to align with and mirror the 6D pose of the human hand. \textbf{(Middle)} With this video demonstration, we train a perceiver-actor agent and \textbf{(Right)} deploy the agent on a real-world robot to demonstrate its ability to learn from AR demonstrations.}
    % \rk{Changes still left: (1) I want the three parts to be a bit symmetric in width. Can we change the left side to also have 4 images that take up the same amount of horizontal space? (2) Can make the images on the left smaller all the font in the figures a bit bigger?}\textcolor{red}{fixed}}
    \label{fig:teaser}

\vspace{-1.3em}
\end{figure}

We introduce \system: a system for collecting robot demonstrations that (1) does not require people to have specialized training, (2) does not require any real robots during data collection, and therefore, (3) enables manipulation of diverse objects with a real robot.
\system\ is a framework in the form of an iOS app that enables users to record a video of themselves manipulating any object. Once the video is captured, \system\ uses the iOS depth sensor to place an AR robot in the scene and uses a motion planner to produce a trajectory where it appears as if the AR robot manipulates the object (Figure~\ref{fig:teaser}).
Manipulating objects and recording a video is so intuitive that users do not need any training to use \system. Our system completely removes the need for a real robot during demonstration collection, allowing data collection to potentially parallelize without being limited by expensive real robots. Finally, since videos can be captured anywhere, \system\ re-situates demonstration collection out of the laboratory; users can take videos anywhere, making it easy to collect demonstrations involving manipulation of diverse objects. Furthermore, unlike collecting visual observations of human activities, our approach uses AR projection during robot demonstration collection to provide constant feedback on the robot's pose and physical constraints in the given environment while performing the task.

% Para 4: Lay out the experiments that were run and what the main findings were.
Our experiments show that \system's AR demonstrations can effectively train a real robot to manipulate real-world objects. We use \system to collect robot demonstrations cross three manipulation tasks (\emph{press}, \emph{pick-up} and \emph{push}) on 9 personalized objects. These personalized objects are uniquely shaped, sized, or textured items designed to meet the specific needs or functionalities of individual users within their personalized environments. We collect and use as few as five effortlessly collected demonstrations to train a behavior cloning agent~\cite{shridhar2022perceiver}.
This trained agent needs to be finetuned for 3,000 iterations (which is equivalent to less than 10 minutes of training) on a dummy real-world task to overcome the sim-to-real gap; Once finetuned, a real robot is capable of manipulating real objects even though that object was only encountered by the AR robot. This AR-trained agent performs comparable to agents trained with real-world tele-operated demonstrations (specifically the PerAct demonstration collection~\cite{shridhar2022perceiver}).

% Para 5: Lay out the user study and the main results there.
We assess \system's usability through a within-subjects user study (N=10). For the user study, participants are asked to provide demonstrations for two standard manipulation tasks: \emph{pick-up} and \emph{stacking}. Besides \system, users collect demonstrations using four alternative methods, including keyboard and mouse \cite{mittal2023orbit}, VR controller \cite{Shridhar2022PerceiverActorAM}, 3D-mouse \cite{mittal2023orbit}, and kinesthetic-teaching \cite{ajaykumar2021designing}. Results suggest \system\ is intuitive, requires no training, and enables quick demonstration production, comparable to kinesthetic teaching and faster than other methods. \system paves the way for democratizing robot training: an estimated 1.36 billion\footnote{Source: \url{https://www.bankmycell.com/blog/number-of-iphone-users}} iPhone users could create personalized manipulation data to train real robots for their household objects.

\begin{figure}[t]
    \centering
    \includegraphics[width=\textwidth]{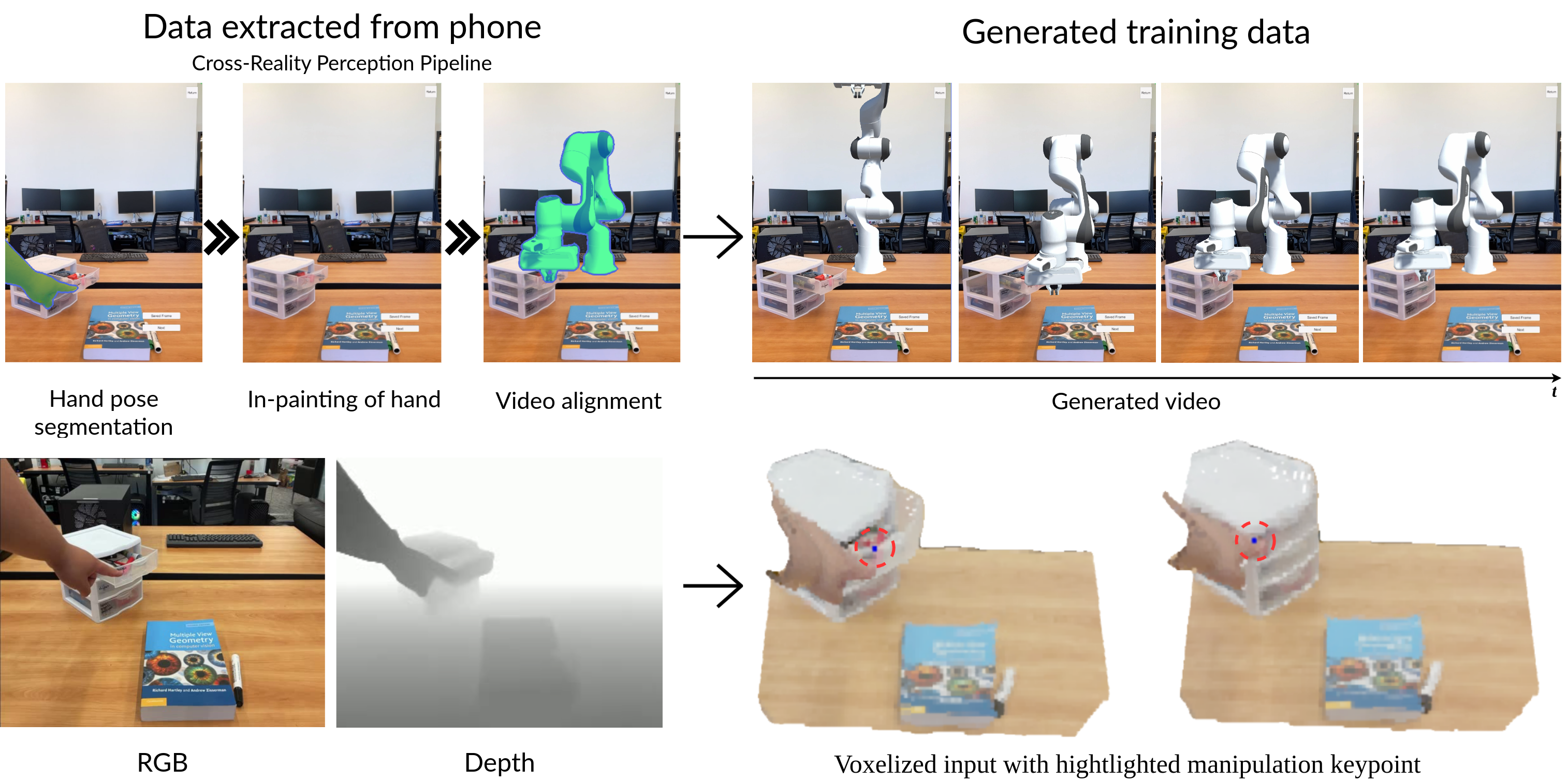}
    \caption{\textbf{\system\ collection process}. \textbf{(Left)} Once the user records themselves manipulating an object, \system\ extracts the following information: 6D hand pose, hand state, RGB frames and depth estimations. We replace the hand with an AR robot, aligning its motions to align its end effector with the hand's. \textbf{(Right)} We create a 3D voxelized representation over time from the extracted information. 
    This 3D representation is used to train a \model~\cite{shridhar2022perceiver} agent. 
    We also use the generated video to train an image-conditioned BC agent \cite{shridhar2022perceiver}.
    % \rk{Looks better. I think we can improve it a bit further: (1) Align all the images so that all the top row are the same height and width and are aligned at the top. Notice how right now the right side has images that are taller and narrower? That doesn't look good. (2) Can we make the images slightly smaller overall and the text font slightly bigger?}\textcolor{red}{fixed the spacing}
    }
    \label{fig:fig2}
    \vspace{-1em}

\end{figure}

%% file: sections/2_related_work.tex
\section{Related work}
We compare our work to existing demonstration collection methods.
 
\textbf{Demonstration collection methods.}
There are several conventional methods available for gathering robot demonstrations.
One popular approach involves kinesthetically controlling the robot to follow a desired trajectory; the generated robot trajectories showcase the behavior required to accomplish a specific task~\cite{argall2009survey}.
Teleoperation techniques~\cite{goldberg1994beyond,lumelsky1993real,hokayem2006bilateral} have also been a popular collection process, using various user interfaces including keyboard and mouse~\cite{kent2017comparison,leeper2012strategies}, a video game controller~\cite{laskey2017comparing}, 3D-mouse~\cite{dragan2012online,shridhar2022perceiver}, mobile phones~\cite{mandlekar2018roboturk}, special purpose master-slave interfaces~\cite{akgun2012novel,liang2017using}, and even virtual reality controllers~\cite{whitney2019comparing,zhang2018deep,lipton2017baxter,perceivertransformer}.
However, all of these methods require a real physical robot to be controlled, bottle-necking demonstration collection by how many robots are available and limited to the laboratories that house these robots.

\textbf{Demonstrations for behavior cloning.}
Recently, robot demonstrations are primarily used as training data for imitation learning, which has pioneered a paradigm shift in robot training.
Offline behavior cloning from robot demonstrations is currently the de-facto imitation learning paradigm~\cite{pomerleau1988alvinn}. 
These demonstrations are collected either in simulation or through human control using a real robot in the real world~\cite{dalal2023imitating,james2022coarse}.
For example, Task and Motion Planning (TAMP) uses expert task planners to create large-scale simulation demonstrations~\cite{dalal2023imitating}.
Meanwhile in the real-world, users employ techniques such as teleportation or vision-based guidance are used to create demonstrations~\cite{zhang2018deep,finn2017one,billard2008robot,wang2023mimicplay}. 
Recent methods have also begun developing specialized hardware to streamline demonstration collection.
For example, a low-cost handheld device featuring a plastic grabber tool outfitted with an RGB-D camera and a servo can control the binary opening and closing of a grabber's fingers~\cite{song2020grasping}. 
By contrast, our real-world data collection approach requires no teleoperation hardware \cite{mandlekar2018roboturk}, no simulators \cite{james2020rlbench}, and most importantly, no real robots \cite{wu2017visual}. All we need is an iPhone camera to record users manipulating objects with their hands. 

%% file: sections/3_method.tex
\section{The \system system}

We introduce AR2-D2, a system for collecting robot demonstrations without requiring a physical robot. In this section, we describe \system's features, its supported data collection procedure, its implementation details. 
% We also conduct an user study contrasting our system against others.

\begin{figure}[t]
    \centering
    \includegraphics[width=\linewidth]{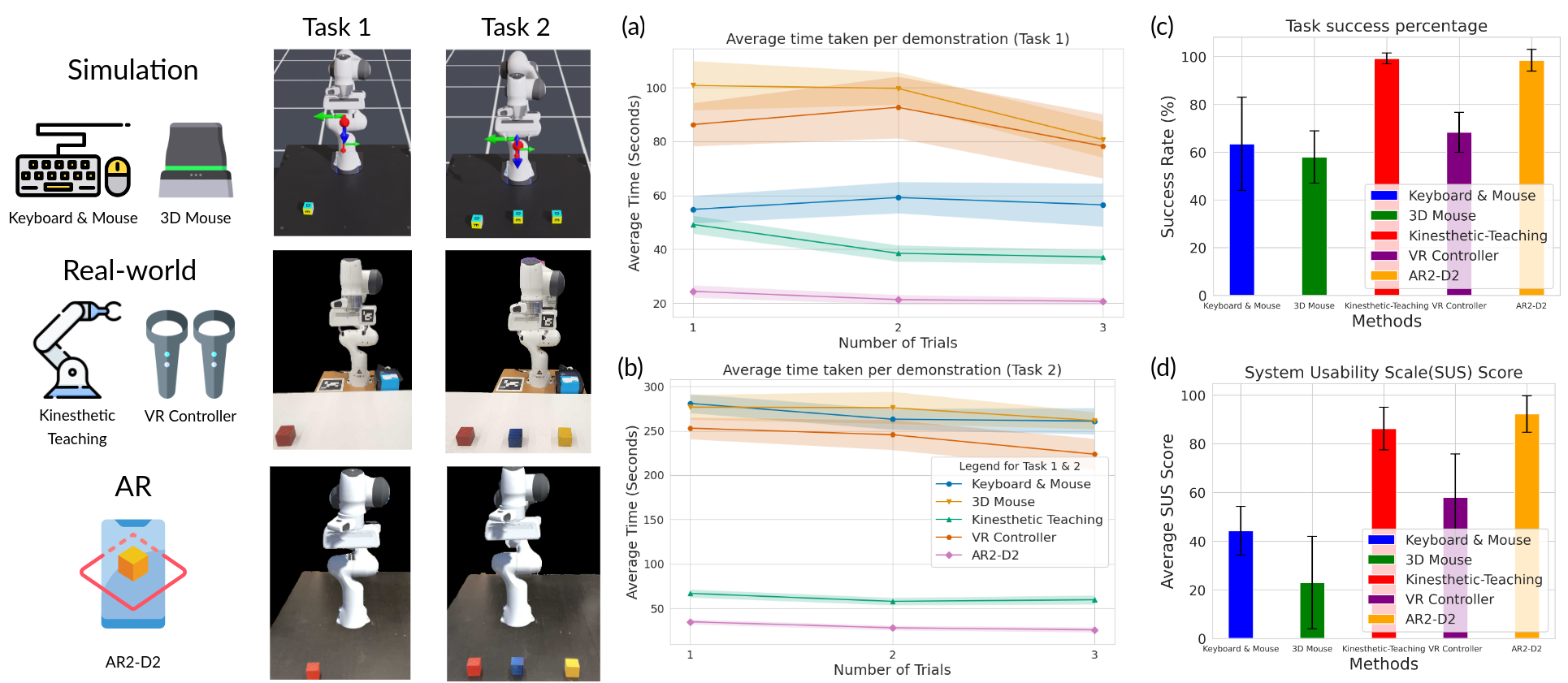}
    \caption{\textbf{Evaluating \system\ with real users.} We conduct an extensive within-subjects user study, comparing \system\ against $4$ alternative  collection techniques: keyboard \& mouse, 3D mouse (6-DoF), kinesthetic teaching, and HTC Hive controller. \textbf{(Left)} The first two techniques control a simulated Franka Panda while the next two a real robot; \system\ manipulates an AR robot in the real world. Participants used these techniques to collect demonstrations for two tasks: (1) pick up and move a cube to a designated location and (2) stack three cubes. 
    \textbf{(Right)} (a, b) We find that participants spend significantly less time (with an average of 22.1 and 29.5 seconds across the two tasks) using our system than others versus the next best (kinesthetic teaching with an average of 41.6 and 61.4 seconds).
    (c, d) We show that participants are able to successfully collect a demonstration with the same rate of success using our system as kinesthetic teaching, both of which have significantly higher success rate versus others.
    % \rk{Looking better. A few more changes: (1) Notice how the word ``real-world'' and ``AR'' appear to be hovering between the first and second row and between the second and third rows, respectively? It looks a bit messy. Can we make it so that they don't go beyond their respective rows?
    % (2) I am thinking more about the arrows that you have on the left and don't feel convinced that they are necessary since they are all the same. Instead, what would be nice if there were visuals for keyboard, for 3D mouse, for VR controller, for AR iPhone/iPad, etc. Can we change those arrows to pictograms for these objects? 
    % (3) I am guessing you haven't had a chance to change too seaborn images yet. Let's do that. Also is the line chart shaded region the std dev or std err? We should be plotting std err. 
    % (4) Ideally I would like all these to be smaller and have more white space between everything.}
    }
    \label{fig:user_study}
    \vspace{-1.2em}

\end{figure}

\subsection{\system\ system features}
\system\ contributes the following features:

\textbf{No need for a physical robot.} In traditional robotics research, obtaining demonstrations often involves operating a physical robot~\cite{finn2017one,billard2008robot,zhang2018deep,song2020grasping}. \system\ presents a new paradigm for collecting demonstrations; it forgoes access to a real robot, enabling users to collect high-quality demonstration data from anywhere with only their mobile devices. 

\textbf{Real-time control of AR robots in the real-world.} \system\ leverages LiDAR sensors, which today are ubiquitous in iPhones and other smartphones to estimate the 3D layout in front of the camera to project an AR robot. LiDAR helps the AR robot obey physical and visual realism. Users can control the AR robot in one of three supported interactions: by pointing at 3D points that the robot's end-effector should move to, by using the iPhone's GUI control, or through AR kinesthetic control (see supplementary material). The projected robot's motions are tightly coupled with the real-world environment, and receives feedback upon collisions with real-world objects.

\textbf{Real-time visualization of task feasibility.} \system\ simplifies the demonstration collection by asking users to specify key-points that the robot end-effector should move to in order to complete a task. Once each key-point is specified, \system\ visualizes the AR robot's motion, moving its end-effector from its current position to the new key-pose. This real-time feedback enables users to assess the feasibility and accuracy of the specified key-point and revise their selections if necessary.

\subsection{Design and implementation}

\system's design and implementation consists of two primary components (Figure \ref{fig:fig2}).
The first component is a phone application that projects AR robots into the real-world, allowing users to interact with physical objects and the AR robot. The second component convert collected videos into a format that can be used to train different behavior cloning agents, which can later be deployed on real robots.

\textbf{The phone application.} 
We designed \system\ as an iOS application, which can be run on an iPhone or iPad. Since modern mobile devices and tablets come equipped with in-built LiDAR, we can effectively place AR robots into the real world. 
The phone application is developed atop the Unity Engine and the AR Foundation kit. The application receives camera sensor data, including camera intrinsic and extrinsic values, and depth directly from the mobile device's built-in functionality. The AR Foundation kit enables the projection of the Franka Panda robot arm's URDF into the physical space. To determine user's 2D hand pose, we utilize Apple's human pose detection algorithm. This, together with the depth map is used to reconstruct the 3D pose of the human hand. By continuously tracking the hand pose at a rate of $30$ frames per second, we can mimic the pose with the AR robot's end-effector.

\textbf{Training data creation.} Given language instructions describing a task (e.g.~``Pick up the plastic bowl''), we hire users to generate demonstrations using AR2-D2. From the user-generated demonstration video, we generate training data that can be used to train and deploy on a real robot. 
To create useful training data, we convert this video into one where it looks like an AR robot manipulated the object. To do so, we segment out and eliminate the human hand using Segment-Anything~\cite{kirillov2023segment}. We fill in the gap left behind by the missing hand with a video in-painting technique known as E2FGVI~\cite{li2022towards}.
Finally, we produce a video with the AR robot arm moving to the key-points identified by the user's hand. This final video processed video makes it appear as if an AR robot arm manipulated the real-world object; it can used as training data for visual-based imitation learning~\cite{young2020visual}.
Additionally, since we have access to the scene's depth estimation, we can also generate a 3D voxelized representation of the scene and use it to train agents like Perceiver-Actor (\model)~\cite{shridhar2022perceiver}.

%% file: sections/4_user_study.tex
\begin{figure}[t]
    \centering
    \includegraphics[width=\linewidth]{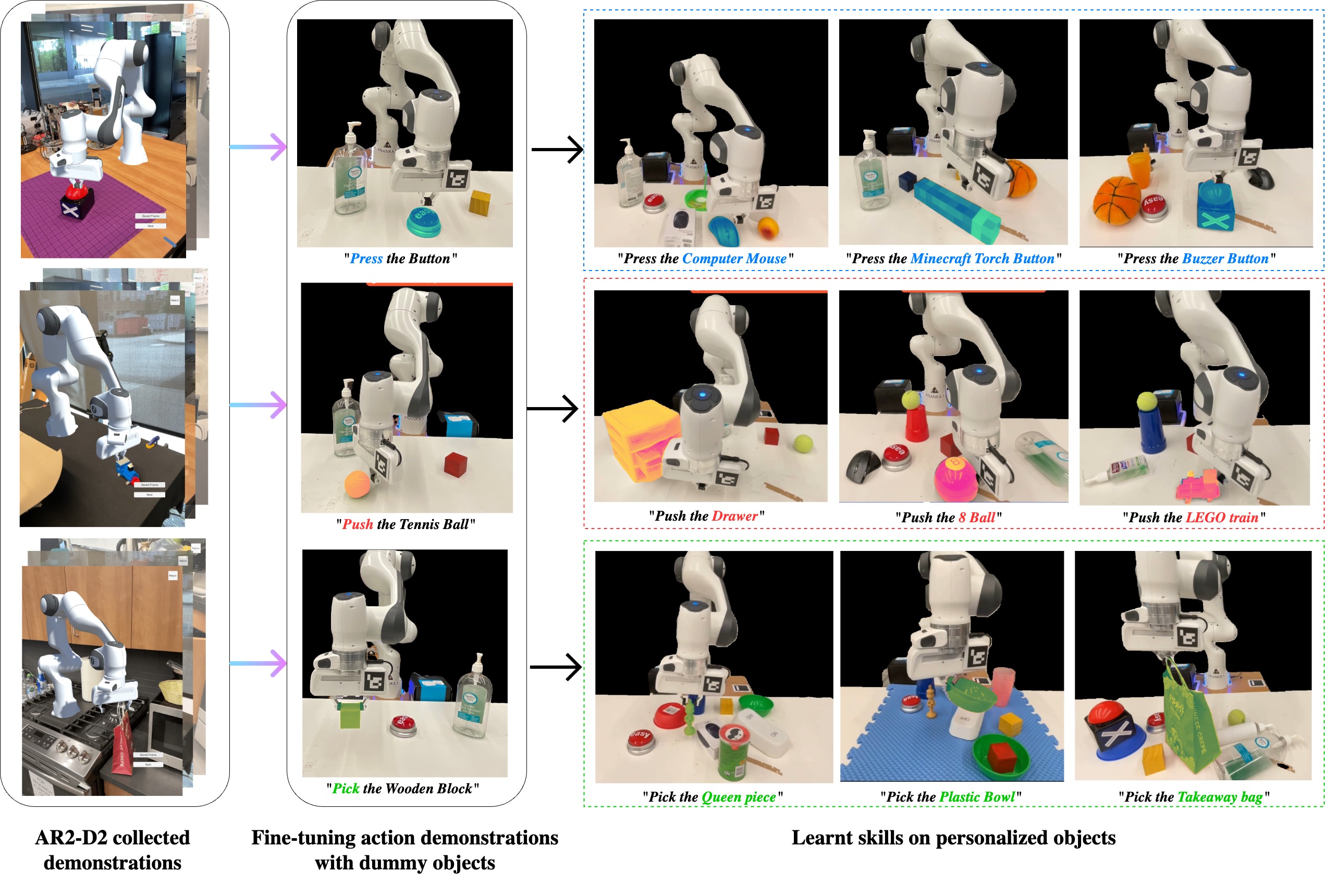}
    \caption{\textbf{Evaluating \system\ data by training a real robot to manipulate real objects}. We employ AR2-D2 as a tool for gathering a diverse array of manipulations encompassing three fundamental actions, involving a wide variety of customized objects. These manipulations range from performing precise actions such as pressing a computer mouse or a Minecraft torch button at specific locations, to pushing small LEGO train toys towards table-sized drawers, and even encompassing the ability to pick up objects varying from chess pieces to takeaway bags. By leveraging a limited number of real-world action demonstrations conducted with random dummy objects and fine-tuning for 3,000 iterations which is equivalent to 10 minutes of training, we have achieved the capacity to apply the PerAct framework in manipulating all these personalized objects with broad generalization.}
    \label{fig:diagnostic_analysis}
    \vspace{-1.2em}

\end{figure}

\section{Evaluating \system\ with real users}
To evaluate \system's efficacy, we conduct an extensive within-subjects user study ($N=10$) across $5$ demonstration collection techniques for $2$ tasks. Participants demonstrated each task $3$ times with each technique, resulting in a total of $300$ collected demonstrations. Participants were locally hired; they were aged between $23$ and $30$.

\noindent\textbf{Baselines collection techniques.}
In order to compare how effectively real participants create demonstrations with \system, we also ask them to use $4$ other baseline collection techniques. Two collection techniques utilize real robots in the real-world and two control simulation robots. 
In simulation, participants control a simulated Franka Panda with either keyboard and mouse or with a 3D Space Mouse.
Using the keyboard and mouse, users can manipulate the 6D end-effector of the simulated robot within the Isaac Sim environment, utilizing ORBIT~\cite{mittal2023orbit}. 
The 3D Space Mouse is a joystick capable of simultaneous translation and rotation along the (x, y, z) axes; it operates within the same environment as the keyboard.
In the real world, participants use kinesthetic teaching or an HTC Vive VR controller.
Kinesthetic teaching allows participants to manipulate a real Franka Panda, using its default zero-gravity feature.
The demonstration collection interface using the HTC Vive VR controller was developed in a recent paper and enables teleoperation of the robot~\cite{perceivertransformer}.

\noindent\textbf{Study protocol.}
Each participant was tasked with collecting demonstrations for two specific tasks: \emph{picking} and a \emph{stacking} (Figure~\ref{fig:user_study}).
Participants were asked to provide demonstrations for each task across $3$ trials, with $3$ attempts allowed per trial. We imposed a time constraint are each trial: $3$ minute limit for the \emph{picking} and a 5 minute limit for \emph{stacking}. After all the data was collected, participants filled out a system usability scale (SUS) survey.

\noindent\textbf{Measured variables.}
We evaluate the different data collection techniques using two metrics. First, we measure average data collection time (in seconds). Lower values are better because it implies that participants are able to collect demonstrations quicker.
Second, we measure task success rate, which calculates the percentage of trials that lead to a successful demonstration.

\noindent\textbf{Results.}
We show that participants using \system\ are both significantly faster (Figure~\ref{fig:diagnostic_analysis} (a, b)) and as likely (Figure~\ref{fig:diagnostic_analysis} (c, d)) to collect a successful demonstration as kinesthetic teaching.
In comparison with kinesthetic teaching, which has an average task completion time of 41.6 and 61.4 seconds for task 1 and 2 respectively, our method exhibits a substantial reduction in time with only 22.1 and 29.5 seconds for both tasks respectively. Furthermore, the t-tests for task 1 and task 2 yielded t-statistics of $t_1 = 6.194$ and $t_2 = 6.199$, with p-values of $p_1 = 7.587 \times 10^{-6}$ and $p_2 = 7.514 \times 10^{-6}$ respectively. Hence, we could confidently say that there is a statistically significant difference between kinesthetic teaching and our approach, with kinesthetic teaching having, on average, significantly longer timings compare to ours.
% Specifically, participants spent and average of $19.47\pm$ and $31.86$ seconds less time per trial for the first and second tasks, respectively.
This concludes that our method is capable of collecting robot demonstrations faster than the traditionally favored kinesthetic teaching. 

We find that participants using \system\ are fast from the get-go (Figure~\ref{fig:user_study}(a, b)).
Participants are consistently faster when collecting demonstrations from the very first trial. This consistency is reflected in the relatively low standard deviation values of $5.75$ and $8.89$ seconds for the two tasks across participants. 
In contrast, the next quickest contender, kinesthetic teaching, exhibits a standard deviation of $9.62$ and $14.02$. Additionally, users have indicated a higher preference for our system in the SUS survey \cite{bangor2008empirical} (Figure \ref{fig:user_study} (d)). Our method garners a similar level of user preference as kinesthetic-teaching, which necessitates a physical robot, with a mere ±6\% difference in SUS scores between the two techniques.

%% file: sections/5_experiments.tex
\section{Evaluating \system\ with a real robot deployment}
\begin{table*}[t]
\begin{center}
    \resizebox{\textwidth}{!}{%z
    \begin{tabular}{l p{0.2pt}ccc p{0.2pt}ccc p{0.2pt}ccc}
    \toprule
     Task& & \multicolumn{3}{c}{Press (\emph{Succ.\%})}&& \multicolumn{3}{c}{Push (\emph{Succ.\%})}&& \multicolumn{3}{c}{Pick up (\emph{Succ.\%})}\\
    \cline{3-5}\cline{7-9}\cline{11-13}
    Personalized object&& Computer mouse & Minecraft torch & Buzzer && LEGO train & 8 ball  & Drawer && Queen piece & Plastic bowl & Takeaway bag \\
    \midrule
    Simulation &&  13.3 & 6.7&30.0 &&13.3 &20.0 &3.3&&3.3&20.0 &16.7 \\
    VR interface (w/o personalized objects)&& 3.3 & 6.7& 16.7&& 13.3 & 10.0 & 3.3&& 0.0& 16.7&13.3 \\
    VR interface (with personalized objects)&& 60.0 & 63.3 & 83.3 && 30.0 & 70.0 & 40.0 && 46.7 & 56.7 & 60.0\\
       AR2-D2 (Ours) &&56.7 & 53.3 & 73.3 && 50.0 &55.7 &23.3 && 46.7 &53.3 & 63.3\\
    \bottomrule
    
    \end{tabular}
    }
\end{center}
\caption{\textbf{Task test results}.We utilized AR2-D2 to collect demonstrations and train BC agents for real robot deployment. Our observations revealed comparable results between our data collection approach and alternative methods. Success rates (mean \%) of the foundational skills tested on personalized objects collected via AR2-D2. For each skill, we evaluated it across ten different sets of distractors with the target object and repeated thrice for consistency. The result has shown that our data collection approach with minimal fine-tuning achieves comparable results to real-world data collected on these personalized objects via PerAct's VR interface with a physical robot.}
\label{tab:press_push_pick_table}
\vspace{-1em}
\end{table*}

With \system, we collect demonstrations and train behavior cloning agents for deployment on a real robot.
Here, we present our experimental setup and three key takeaways. First, we validate that \system\ demonstrations can train a real robot to manipulate personalized objects without access to a physical robot. Second, the agent trained using \system's demonstrations perform on par with training on real robot demonstrations. Third, \system's demonstrations can enable learning policies from both image as well as voxelized inputs.

\noindent\textbf{\system\ demonstration collection.}
We collect \system\ demonstrations on a set of personalized objects, and demonstrate that a policy trained on this data executes on a real Franka Panda robot. We gather demonstrations centered around three common robotics tasks: {\{\emph{press, push, pick up}\}. For each task, we collect five demonstrations using three different objects, which vary in color, size, geometry, texture, and even functionality (see Figure~\ref{fig:diagnostic_analysis}).

\noindent\textbf{Behaviour cloning.} We use Perciver-Actor (\model)~\cite{shridhar2022perceiver} to train a transformer-based language-guided behavior cloning policy. \model\ takes a 3D voxel observation and a language goal $(v, l)$ as input and produces discretized outputs for translation, rotation, and gripper state of the end-effector. These outputs, coupled with a motion planner, enable the execution of the task specified by the language goal.

\noindent\textbf{Training procedure.} Following existing work~\cite{shridhar2022perceiver}, we train an individual agent for each task. We train an agent for $30$k iterations per set of demonstrations. We then freeze the backbone of the \model\ architecture and finetune the rest using the set of VR (using HTC Hive) demonstrations on dummy objects. This fine-tuning process spans an additional $3$k iterations, equivalent to approximately $10$ minutes of wall clock training. Fine-tuning allows us to close the domain gap resulting from differences in depth cameras between the Kinect V2 used by \model\ and the iPhone/iPad depth camera used by \system.

\begin{wraptable}{r}{0.51\textwidth}
  %\begin{table}[h]
%   \hspace*{-1em}
%   \vspace{-0.5cm}
  \vspace{-1.3em}
  \setlength\tabcolsep{2.3pt}
  \centering
  \scriptsize
\begin{tabular}{lcc} 
\toprule
Task          & 2D data (Image-BC)~ & 3D data (PerAct)  \\ 
\midrule
Press the buzzer from the side & 0.00\%         & 40.00\%    \\
Pick up the queen piece    & 6.67\%         & 33.34\%           \\
Press the computer mouse    & 6.67\%        & 40.00\%           \\
\bottomrule
\end{tabular}
%   \vspace{-0.5em}
    \caption{\textbf{Training with Different Data Modalities.} AR2-D2 is capable of offering diverse data modalities to facilitate training BC models, such as Image-BC for 2D data and PerAct for 3D data. We assess these disparate data modalities, gathered via AR2-D2, across three distinct tasks using Image-BC for 2D data and PerAct for 3D data, conducted without any form of fine-tuning} % task, success \% vs. $\#$ of demonstrations. 
  \vspace{-2em}
  \label{table:table2}
\end{wraptable}

\noindent\textbf{Finetuning demonstration collection.}
Finetuning demonstrations are collected using the VR interface from PerAct~\cite{shridhar2022perceiver}. It involves using a VR handset to guide the real-robot to desired end-effector poses. 
We collect $5$ demonstrations for each task using three dummy objects: {\{\emph{red button, yellow block, tennis ball}\}, corresponding to the three tasks, respectively. These specific objects are only used for finetuning and not used during testing.
We also ablate the agent's performance without finetuning.

\begin{wrapfigure}{r}{0.4\textwidth}
%   \vspace{-1.3em}
%   \vspace{-2.3em}
%   \vspace{-4.3em}
  
  \begin{center}
    \vspace{-0.8cm}
    % \hspace{2.5cm}
    \includegraphics[width=0.40\textwidth]{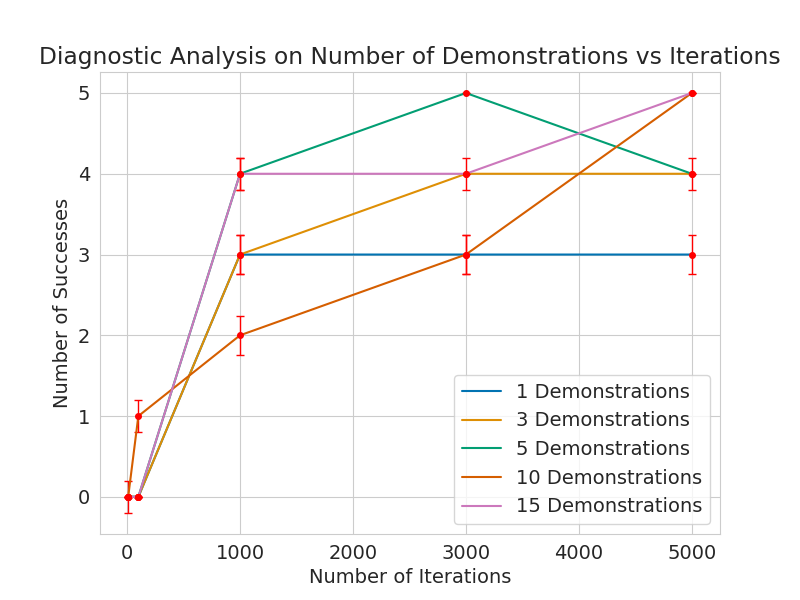}

%   \vspace{-1.2em}
  \vspace{-0.05cm}
  \caption{\textbf{Analysis on Fine-tuning.} We conducted a diagnostic analysis to determine the optimal number of iterations and demonstrations required. By varying the number of demonstrations and iterations for fine-tuning, we found that using 5 demonstrations and 3,000 iterations yielded the best results.}
  \label{fig:ablations}
  \end{center}
  \vspace{-2em}
%   \vspace{-1.7em}
%   \vspace{-0.5em}
\end{wrapfigure}
\vspace{-0.05cm}

\noindent\textbf{Testing procedure.} 
% During testing, we exclusively evaluate the pre-trained   AR2-D2-PerAct model on the distinctly different objects from the   AR2-D2-enabled real-world demonstrations. To ensure consistency, we evaluate across ten different sets of distractors, and repeat three times for each setting to obtain the average performance. The rationale behind this training strategy is to demonstrate that with   AR2-D2, we can collect personalized objects in custom environments, enabling training and facilitating the model's ability to generalize and manipulate these objects. The fine-tuning stage serves as a grounding process, allowing the model to adapt to the existing environment. We also evaluated directly using   AR2-D2 demonstrations which was trained on PerAct for 30,000 iterations and with no fine-tuning in our ablation study.
We evaluate the trained policies' ability to manipulate personalized objects in the real world. The personalized objects are comprised of distinctly different objects from the \system-enabled real-world demonstrations. Each test environment is infected with ten different distractor objects. We repeat run inference three times for each environment setup and average their performance.

\textbf{Baseline collection techniques.} We compare \system's demonstrations against two alternative techniques: real-world and simulation data collection. 
We finetune all methods using the same set of finetuning demonstrations on dummy objects. 
\textit{Real world} data collection uses a VR controller interface to capture the training demonstrations~\cite{shridhar2022perceiver}. Real-world demonstrations are collected with and without the personalized objects (see Table \ref{tab:press_push_pick_table}). \textit{Simulation} demonstrations use RLBench and its key-frame point extraction technique, accompanied by motion planning to generate each demonstration~\cite{james2020rlbench}. We implemented domain randomization to introduce texture variations, aiding in the transfer from simulation to the real world.

\subsection{Results}

Table~\ref{tab:press_push_pick_table} reports success rates of all the nine personalized objects across three tasks with demonstrations from real-world, from simulation, and from \system. 

\textbf{\system demonstrations yields useful representation for training a real-robot. } In general, \system's data outperforms policies trained using simulation of real-world (without personalized objects). 
In fact, in one case, we outperform \model's real world data collection (without personalized objects) by a large margin of $53.4\%$. These findings highlight the significance of our approach, which facilitates access to collecting demonstrations with such personalized objects which might not be available in the laboratory that houses the robot. This capability to produce training data with personalized objects is particularly important since behavior cloning agents perform better when their training exposures them to the  objects they are expected to manipulate. 

\textbf{\system\ demonstrations train policies as accurately as demonstrations collected from real robots.} Referencing Table~\ref{tab:press_push_pick_table}, it is evident that even when real-world data collection is trained with personalized objects during the demonstration data collection phase, our method delivers comparable results. Remarkably, our system's data even surpasses the real-robot collection data in tasks such as pushing the LEGO train and picking up the paper bag. While for the remaining personalized objects, our approach maintains a $\leq 14.3 \% $ gap across the three foundational skills. The t-test results, with a calculated \emph{t-value} of 0.547 and a \emph{p-value} of 0.592, indicating that there is no statistical significance in the observed difference between the two methods.

% \begin{wraptable}{r}{0.1\textwidth}
%   %\begin{table}[h]
% %   \hspace*{-1em}
% %   \vspace{-0.5cm}
%   \vspace{-1.3em}
%   \setlength\tabcolsep{2.3pt}
%   \centering
%   \scriptsize
% \begin{tabular}{lcc} 
% \toprule
% Task          & 2D data (Image-BC)~ & 3D data (PerAct)  \\ 
% \midrule
% Press the buzzer from the side & 0.00\%         & 40.00\%    \\
% Pick up the queen piece    & 6.67\%         & 33.34\%           \\
% Press the computer mouse    & 6.67\%        & 40.00\%           \\
% \bottomrule
% \end{tabular}
% %   \vspace{-0.5em}
%     \caption{\textbf{Training with Different Data Modalities.} AR2-D2 is capable of offering diverse data modalities to facilitate training BC models, such as Image-BC for 2D data and PerAct for 3D data. We assess these disparate data modalities, gathered via AR2-D2, across three distinct tasks using Image-BC for 2D data and PerAct for 3D data, conducted without any form of fine-tuning} % task, success \% vs. $\#$ of demonstrations. 
%   % \vspace{-2em}
%   \label{table:table2}
% \end{wraptable}

\subsection{Ablations}
\noindent\textbf{Analysis on Fine-tuning.} We investigate how many finetuning demonstrations ($\{1, 3, 5, 10, 15\}$) on dummy objects and how many training iterations ($\{0, 1000, 2000, 3000, 4000, 5000\}$) are required to maximize the agent's performance.
These ablations pretrain the policy using $5$ \system\ demonstrations of the ``mouse" pressing task trained for $30$k training iterations.
Each ablation is tested on $1$ real-world scene with the computer mouse but we evaluated it across $5$ trials with varying target object poses and placement.
We find that $5$ fine-tuning demonstrations trained for $3$k iterations (equivalent to 10 minutes of training) yields the most effective outcome (see Figure \ref{fig:ablations}).

\noindent\textbf{Training with voxelized inputs is better than using 2D inputs.} \system\ demonstrations store 2D image and 3D depth data, facilitating training of image-based behavior cloning (Image-BC) and 3D voxelized methods (\model~\cite{shridhar2022perceiver}).
With fixed camera calibration offset and no finetuning during training, 3D-input agents outperform 2D counterparts (refer to Table~\ref{table:table2} and supplementary).

%% file: sections/6_conclusion.tex
\section{Limitations and Conclusion}

\noindent\textbf{Limitations.}
Our research does present certain limitations. Firstly, due to the inherent characteristics of our method, it proves challenging to validate experimental outcomes via simulation. Consequently, the verification relies on real-world assessments, which, despite our extensive multi-trial evaluations using varied layouts, cannot completely encompass all conceivable scenarios. Secondly, while our user-study participant count mirrors the standards set by RoboTurk \cite{mandlekar2018roboturk}, we acknowledge that a larger participant pool might have enhanced the statistical significance of the performance results across various methods. Lastly, due to the disparity between the camera sensors and the domain gap, there is still a need for fine-tuning to match the performance of real data. Nevertheless, future work can explore better approaches to further bridge this domain gap either through better data augmentation techniques or hardware such as Apple's AR head-mounted display.

\noindent\textbf{Conclusion.}
We present AR2-D2, an intuitive robot demonstration collection system that enables the collection of quality robot demonstrations for diverse objects without the need for any real robots or the need to train people before use. Our results highlight the effectiveness of this approach, showing that as few as five AR demonstrations suffice to train a real-world robot to manipulate personalized objects. Our extensive real-world experiments further confirmed that AR2-D2's AR data is on par with training using real-world demonstrations. Moreover, through our comprehensive user-study, it revealed that users found our method intuitive and easy to use, requiring no prior training, setting it apart from traditional collection methods. Finally, AR2-D2 paves the way towards democratizing robot training by enabling any individual to gather significant robot training data for manipulating their personalized objects at any place and time.

%% file: main.bbl
\begin{thebibliography}{41}
\providecommand{\natexlab}[1]{#1}
\providecommand{\url}[1]{\texttt{#1}}
\expandafter\ifx\csname urlstyle\endcsname\relax
  \providecommand{\doi}[1]{doi: #1}\else
  \providecommand{\doi}{doi: \begingroup \urlstyle{rm}\Url}\fi

\bibitem[Radford et~al.(2021)Radford, Kim, Hallacy, Ramesh, Goh, Agarwal,
  Sastry, Askell, Mishkin, Clark, et~al.]{radford2021learning}
A.~Radford, J.~W. Kim, C.~Hallacy, A.~Ramesh, G.~Goh, S.~Agarwal, G.~Sastry,
  A.~Askell, P.~Mishkin, J.~Clark, et~al.
\newblock Learning transferable visual models from natural language
  supervision.
\newblock In \emph{International conference on machine learning}, pages
  8748--8763. PMLR, 2021.

\bibitem[Ramesh et~al.(2022)Ramesh, Dhariwal, Nichol, Chu, and
  Chen]{ramesh2022hierarchical}
A.~Ramesh, P.~Dhariwal, A.~Nichol, C.~Chu, and M.~Chen.
\newblock Hierarchical text-conditional image generation with clip latents.
\newblock \emph{arXiv preprint arXiv:2204.06125}, 2022.

\bibitem[Duan et~al.(2022)Duan, Yu, Tan, Zhu, and Tan]{duan2022survey}
J.~Duan, S.~Yu, H.~L. Tan, H.~Zhu, and C.~Tan.
\newblock A survey of embodied ai: From simulators to research tasks.
\newblock \emph{IEEE Transactions on Emerging Topics in Computational
  Intelligence}, 6\penalty0 (2):\penalty0 230--244, 2022.

\bibitem[Dasari et~al.(2019)Dasari, Ebert, Tian, Nair, Bucher, Schmeckpeper,
  Singh, Levine, and Finn]{dasari2019robonet}
S.~Dasari, F.~Ebert, S.~Tian, S.~Nair, B.~Bucher, K.~Schmeckpeper, S.~Singh,
  S.~Levine, and C.~Finn.
\newblock Robonet: Large-scale multi-robot learning.
\newblock \emph{arXiv preprint arXiv:1910.11215}, 2019.

\bibitem[Ebert et~al.(2021)Ebert, Yang, Schmeckpeper, Bucher, Georgakis,
  Daniilidis, Finn, and Levine]{ebert2021bridge}
F.~Ebert, Y.~Yang, K.~Schmeckpeper, B.~Bucher, G.~Georgakis, K.~Daniilidis,
  C.~Finn, and S.~Levine.
\newblock Bridge data: Boosting generalization of robotic skills with
  cross-domain datasets.
\newblock \emph{arXiv preprint arXiv:2109.13396}, 2021.

\bibitem[Shridhar et~al.(2022)Shridhar, Manuelli, and
  Fox]{shridhar2022perceiver}
M.~Shridhar, L.~Manuelli, and D.~Fox.
\newblock Perceiver-actor: A multi-task transformer for robotic manipulation.
\newblock \emph{arXiv preprint arXiv:2209.05451}, 2022.

\bibitem[Wang et~al.(2023)Wang, Fan, Sun, Zhang, Fei-Fei, Xu, Zhu, and
  Anandkumar]{wang2023mimicplay}
C.~Wang, L.~Fan, J.~Sun, R.~Zhang, L.~Fei-Fei, D.~Xu, Y.~Zhu, and
  A.~Anandkumar.
\newblock Mimicplay: Long-horizon imitation learning by watching human play.
\newblock \emph{arXiv preprint arXiv:2302.12422}, 2023.

\bibitem[Ahn et~al.(2022)Ahn, Brohan, Brown, Chebotar, Cortes, David, Finn,
  Gopalakrishnan, Hausman, Herzog, et~al.]{ahn2022can}
M.~Ahn, A.~Brohan, N.~Brown, Y.~Chebotar, O.~Cortes, B.~David, C.~Finn,
  K.~Gopalakrishnan, K.~Hausman, A.~Herzog, et~al.
\newblock Do as i can, not as i say: Grounding language in robotic affordances.
\newblock \emph{arXiv preprint arXiv:2204.01691}, 2022.

\bibitem[Jiang et~al.(2022)Jiang, Gupta, Zhang, Wang, Dou, Chen, Fei-Fei,
  Anandkumar, Zhu, and Fan]{jiang2022vima}
Y.~Jiang, A.~Gupta, Z.~Zhang, G.~Wang, Y.~Dou, Y.~Chen, L.~Fei-Fei,
  A.~Anandkumar, Y.~Zhu, and L.~Fan.
\newblock Vima: General robot manipulation with multimodal prompts.
\newblock \emph{arXiv preprint arXiv:2210.03094}, 2022.

\bibitem[Brohan et~al.(2022)Brohan, Brown, Carbajal, Chebotar, Dabis, Finn,
  Gopalakrishnan, Hausman, Herzog, Hsu, et~al.]{brohan2022rt}
A.~Brohan, N.~Brown, J.~Carbajal, Y.~Chebotar, J.~Dabis, C.~Finn,
  K.~Gopalakrishnan, K.~Hausman, A.~Herzog, J.~Hsu, et~al.
\newblock Rt-1: Robotics transformer for real-world control at scale.
\newblock \emph{arXiv preprint arXiv:2212.06817}, 2022.

\bibitem[Argall et~al.(2009)Argall, Chernova, Veloso, and
  Browning]{argall2009survey}
B.~D. Argall, S.~Chernova, M.~Veloso, and B.~Browning.
\newblock A survey of robot learning from demonstration.
\newblock \emph{Robotics and autonomous systems}, 57\penalty0 (5):\penalty0
  469--483, 2009.

\bibitem[Osentoski et~al.(2010)Osentoski, Crick, Jay, and
  Jenkins]{osentoski2010crowdsourcing}
S.~Osentoski, C.~Crick, G.~Jay, and O.~C. Jenkins.
\newblock Crowdsourcing for closed loop control.
\newblock In \emph{Proc. of the NIPS Workshop on Computational Social Science
  and the Wisdom of Crowds, NIPS}, pages 4--7, 2010.

\bibitem[Kent et~al.(2017)Kent, Saldanha, and Chernova]{kent2017comparison}
D.~Kent, C.~Saldanha, and S.~Chernova.
\newblock A comparison of remote robot teleoperation interfaces for general
  object manipulation.
\newblock In \emph{Proceedings of the 2017 ACM/IEEE International Conference on
  Human-Robot Interaction}, pages 371--379, 2017.

\bibitem[Leeper et~al.(2012)Leeper, Hsiao, Ciocarlie, Takayama, and
  Gossow]{leeper2012strategies}
A.~E. Leeper, K.~Hsiao, M.~Ciocarlie, L.~Takayama, and D.~Gossow.
\newblock Strategies for human-in-the-loop robotic grasping.
\newblock In \emph{Proceedings of the seventh annual ACM/IEEE international
  conference on Human-Robot Interaction}, pages 1--8, 2012.

\bibitem[Laskey et~al.(2017)Laskey, Chuck, Lee, Mahler, Krishnan, Jamieson,
  Dragan, and Goldberg]{laskey2017comparing}
M.~Laskey, C.~Chuck, J.~Lee, J.~Mahler, S.~Krishnan, K.~Jamieson, A.~Dragan,
  and K.~Goldberg.
\newblock Comparing human-centric and robot-centric sampling for robot deep
  learning from demonstrations.
\newblock In \emph{2017 IEEE International Conference on Robotics and
  Automation (ICRA)}, pages 358--365. IEEE, 2017.

\bibitem[Dragan and Srinivasa(2012)]{dragan2012online}
A.~D. Dragan and S.~S. Srinivasa.
\newblock Online customization of teleoperation interfaces.
\newblock In \emph{2012 IEEE RO-MAN: The 21st IEEE International Symposium on
  Robot and Human Interactive Communication}, pages 919--924. IEEE, 2012.

\bibitem[Akg{\"u}n et~al.(2012)Akg{\"u}n, Subramanian, and
  Thomaz]{akgun2012novel}
B.~Akg{\"u}n, K.~Subramanian, and A.~L. Thomaz.
\newblock Novel interaction strategies for learning from teleoperation.
\newblock In \emph{AAAI Fall Symposium: Robots Learning Interactively from
  Human Teachers}, volume~12, page~07, 2012.

\bibitem[Liang et~al.(2017)Liang, Mahler, Laskey, Li, and
  Goldberg]{liang2017using}
J.~Liang, J.~Mahler, M.~Laskey, P.~Li, and K.~Goldberg.
\newblock Using dvrk teleoperation to facilitate deep learning of automation
  tasks for an industrial robot.
\newblock In \emph{2017 13th IEEE Conference on Automation Science and
  Engineering (CASE)}, pages 1--8. IEEE, 2017.

\bibitem[Whitney et~al.(2019)Whitney, Rosen, Phillips, Konidaris, and
  Tellex]{whitney2019comparing}
D.~Whitney, E.~Rosen, E.~Phillips, G.~Konidaris, and S.~Tellex.
\newblock Comparing robot grasping teleoperation across desktop and virtual
  reality with ros reality.
\newblock In \emph{Robotics Research: The 18th International Symposium ISRR},
  pages 335--350. Springer, 2019.

\bibitem[Zhang et~al.(2018)Zhang, McCarthy, Jow, Lee, Chen, Goldberg, and
  Abbeel]{zhang2018deep}
T.~Zhang, Z.~McCarthy, O.~Jow, D.~Lee, X.~Chen, K.~Goldberg, and P.~Abbeel.
\newblock Deep imitation learning for complex manipulation tasks from virtual
  reality teleoperation.
\newblock In \emph{2018 IEEE International Conference on Robotics and
  Automation (ICRA)}, pages 5628--5635. IEEE, 2018.

\bibitem[Lipton et~al.(2017)Lipton, Fay, and Rus]{lipton2017baxter}
J.~I. Lipton, A.~J. Fay, and D.~Rus.
\newblock Baxter's homunculus: Virtual reality spaces for teleoperation in
  manufacturing.
\newblock \emph{IEEE Robotics and Automation Letters}, 3\penalty0 (1):\penalty0
  179--186, 2017.

\bibitem[Mittal et~al.(2023)Mittal, Yu, Yu, Liu, Rudin, Hoeller, Yuan, Singh,
  Guo, Mazhar, Mandlekar, Babich, State, Hutter, and Garg]{mittal2023orbit}
M.~Mittal, C.~Yu, Q.~Yu, J.~Liu, N.~Rudin, D.~Hoeller, J.~L. Yuan, R.~Singh,
  Y.~Guo, H.~Mazhar, A.~Mandlekar, B.~Babich, G.~State, M.~Hutter, and A.~Garg.
\newblock Orbit: A unified simulation framework for interactive robot learning
  environments.
\newblock \emph{IEEE Robotics and Automation Letters}, pages 1--8, 2023.
\newblock \doi{10.1109/LRA.2023.3270034}.

\bibitem[Shridhar et~al.(2022)Shridhar, Manuelli, and
  Fox]{Shridhar2022PerceiverActorAM}
M.~Shridhar, L.~Manuelli, and D.~Fox.
\newblock Perceiver-actor: A multi-task transformer for robotic manipulation.
\newblock \emph{ArXiv}, abs/2209.05451, 2022.

\bibitem[Ajaykumar et~al.(2021)Ajaykumar, Stiber, and
  Huang]{ajaykumar2021designing}
G.~Ajaykumar, M.~Stiber, and C.-M. Huang.
\newblock Designing user-centric programming aids for kinesthetic teaching of
  collaborative robots.
\newblock \emph{Robotics and Autonomous Systems}, 145:\penalty0 103845, 2021.

\bibitem[Goldberg et~al.(1994)Goldberg, Mascha, Gentner, Tossman, Rothenberg,
  Sutter, and Wiegley]{goldberg1994beyond}
K.~Goldberg, M.~Mascha, S.~Gentner, J.~Tossman, N.~Rothenberg, C.~Sutter, and
  J.~Wiegley.
\newblock Beyond the web: Excavating the real world via mosaic.
\newblock In \emph{Second International WWW Conference}, pages 1--12, 1994.

\bibitem[Lumelsky and Cheung(1993)]{lumelsky1993real}
V.~J. Lumelsky and E.~Cheung.
\newblock Real-time collision avoidance in teleoperated whole-sensitive robot
  arm manipulators.
\newblock \emph{IEEE Transactions on Systems, Man, and Cybernetics},
  23\penalty0 (1):\penalty0 194--203, 1993.

\bibitem[Hokayem and Spong(2006)]{hokayem2006bilateral}
P.~F. Hokayem and M.~W. Spong.
\newblock Bilateral teleoperation: An historical survey.
\newblock \emph{Automatica}, 42\penalty0 (12):\penalty0 2035--2057, 2006.

\bibitem[Mandlekar et~al.(2018)Mandlekar, Zhu, Garg, Booher, Spero, Tung, Gao,
  Emmons, Gupta, Orbay, et~al.]{mandlekar2018roboturk}
A.~Mandlekar, Y.~Zhu, A.~Garg, J.~Booher, M.~Spero, A.~Tung, J.~Gao, J.~Emmons,
  A.~Gupta, E.~Orbay, et~al.
\newblock Roboturk: A crowdsourcing platform for robotic skill learning through
  imitation.
\newblock In \emph{Conference on Robot Learning}, pages 879--893. PMLR, 2018.

\bibitem[Jaegle et~al.(2021)Jaegle, Gimeno, Brock, Zisserman, Vinyals, and
  Carreira]{perceivertransformer}
A.~Jaegle, F.~Gimeno, A.~Brock, A.~Zisserman, O.~Vinyals, and J.~Carreira.
\newblock Perceiver: General perception with iterative attention, 2021.
\newblock URL \url{https://arxiv.org/abs/2103.03206}.

\bibitem[Pomerleau(1988)]{pomerleau1988alvinn}
D.~A. Pomerleau.
\newblock Alvinn: An autonomous land vehicle in a neural network.
\newblock \emph{Advances in neural information processing systems}, 1, 1988.

\bibitem[Dalal et~al.(2023)Dalal, Mandlekar, Garrett, Handa, Salakhutdinov, and
  Fox]{dalal2023imitating}
M.~Dalal, A.~Mandlekar, C.~Garrett, A.~Handa, R.~Salakhutdinov, and D.~Fox.
\newblock Imitating task and motion planning with visuomotor transformers,
  2023.

\bibitem[James et~al.(2022)James, Wada, Laidlow, and Davison]{james2022coarse}
S.~James, K.~Wada, T.~Laidlow, and A.~J. Davison.
\newblock Coarse-to-fine q-attention: Efficient learning for visual robotic
  manipulation via discretisation.
\newblock In \emph{Proceedings of the IEEE/CVF Conference on Computer Vision
  and Pattern Recognition}, pages 13739--13748, 2022.

\bibitem[Finn et~al.(2017)Finn, Yu, Zhang, Abbeel, and Levine]{finn2017one}
C.~Finn, T.~Yu, T.~Zhang, P.~Abbeel, and S.~Levine.
\newblock One-shot visual imitation learning via meta-learning.
\newblock In \emph{Conference on robot learning}, pages 357--368. PMLR, 2017.

\bibitem[Billard et~al.(2008)Billard, Calinon, Dillmann, and
  Schaal]{billard2008robot}
A.~Billard, S.~Calinon, R.~Dillmann, and S.~Schaal.
\newblock Robot programming by demonstration.
\newblock In \emph{Springer handbook of robotics}, pages 1371--1394. Springer,
  2008.

\bibitem[Song et~al.(2020)Song, Zeng, Lee, and Funkhouser]{song2020grasping}
S.~Song, A.~Zeng, J.~Lee, and T.~Funkhouser.
\newblock Grasping in the wild: Learning 6dof closed-loop grasping from
  low-cost demonstrations.
\newblock \emph{IEEE Robotics and Automation Letters}, 5\penalty0 (3):\penalty0
  4978--4985, 2020.

\bibitem[James et~al.(2020)James, Ma, Arrojo, and Davison]{james2020rlbench}
S.~James, Z.~Ma, D.~R. Arrojo, and A.~J. Davison.
\newblock Rlbench: The robot learning benchmark \& learning environment.
\newblock \emph{IEEE Robotics and Automation Letters}, 5\penalty0 (2):\penalty0
  3019--3026, 2020.

\bibitem[Wu et~al.(2017)Wu, Teney, Wang, Shen, Dick, and van~den
  Hengel]{wu2017visual}
Q.~Wu, D.~Teney, P.~Wang, C.~Shen, A.~Dick, and A.~van~den Hengel.
\newblock Visual question answering: A survey of methods and datasets.
\newblock \emph{Computer Vision and Image Understanding}, 163:\penalty0 21--40,
  2017.

\bibitem[Kirillov et~al.(2023)Kirillov, Mintun, Ravi, Mao, Rolland, Gustafson,
  Xiao, Whitehead, Berg, Lo, et~al.]{kirillov2023segment}
A.~Kirillov, E.~Mintun, N.~Ravi, H.~Mao, C.~Rolland, L.~Gustafson, T.~Xiao,
  S.~Whitehead, A.~C. Berg, W.-Y. Lo, et~al.
\newblock Segment anything.
\newblock \emph{arXiv preprint arXiv:2304.02643}, 2023.

\bibitem[Li et~al.(2022)Li, Lu, Qin, Guo, and Cheng]{li2022towards}
Z.~Li, C.-Z. Lu, J.~Qin, C.-L. Guo, and M.-M. Cheng.
\newblock Towards an end-to-end framework for flow-guided video inpainting.
\newblock In \emph{Proceedings of the IEEE/CVF Conference on Computer Vision
  and Pattern Recognition}, pages 17562--17571, 2022.

\bibitem[Young et~al.(2020)Young, Gandhi, Tulsiani, Gupta, Abbeel, and
  Pinto]{young2020visual}
S.~Young, D.~Gandhi, S.~Tulsiani, A.~Gupta, P.~Abbeel, and L.~Pinto.
\newblock Visual imitation made easy, 2020.

\bibitem[Bangor et~al.(2008)Bangor, Kortum, and Miller]{bangor2008empirical}
A.~Bangor, P.~T. Kortum, and J.~T. Miller.
\newblock An empirical evaluation of the system usability scale.
\newblock \emph{Intl. Journal of Human--Computer Interaction}, 24\penalty0
  (6):\penalty0 574--594, 2008.

\end{thebibliography}
